\documentclass[10pt,twocolumn,letterpaper]{article}

\usepackage{wacv}
\usepackage{times}
\usepackage{epsfig}
\usepackage{graphicx}
\usepackage{amsmath}
\usepackage{amssymb}
\usepackage{color}

\usepackage{color}
\usepackage{breqn} 
\usepackage{tabularx}
\usepackage{booktabs} 
\usepackage{multirow}

\usepackage{epstopdf}
\usepackage{footnote}
\makesavenoteenv{tabular}
\makesavenoteenv{table}

\DeclareMathOperator*{\argmax}{argmax}

\newcommand{\com}[1] {}

\newcommand*{\affaddr}[1]{#1} 
\newcommand*{\affmark}[1][*]{\textsuperscript{#1}}

\wacvfinalcopy 

\def\wacvPaperID{102} 

\ifwacvfinal\fi
\begin{document}

\title{A Multi-view RGB-D Approach for Human Pose Estimation in Operating Rooms}


\author{%
Abdolrahim Kadkhodamohammadi\affmark[1], Afshin Gangi\affmark[1,2], Michel de Mathelin\affmark[1], Nicolas Padoy\affmark[1]\\
\affaddr{\affmark[1]ICube, University of Strasbourg, CNRS, IHU Strasbourg, France}\\
\affaddr{\affmark[2]Radiology Department, University Hospital of Strasbourg, France}\\
{\tt\small \{kadkhodamohammad, gangi, demathelin, npadoy\}@unistra.fr}\\
}

%

\maketitle
\ifwacvfinal\thispagestyle{empty}\fi

\begin{abstract}
  Many approaches have been proposed for human pose estimation in single and multi-view RGB images. However, some environments, such as the operating room, are still very challenging for state-of-the-art RGB methods.  
  In this paper, we propose an approach for multi-view 3D human pose estimation from RGB-D images and demonstrate the benefits of using the additional depth channel for pose refinement beyond its use for the generation of improved features. The proposed method permits the joint detection and estimation of the poses without knowing a priori the number of persons present in the scene. We evaluate this approach on a novel multi-view RGB-D dataset acquired during live surgeries and annotated with ground truth 3D poses.    
\end{abstract}

\section{Introduction}

Recovering the configuration of human body parts, which is also referred to as Human Pose Estimation (HPE), can benefit a wide variety of applications such as video surveillance and behavior analysis~\cite{gowsikhaa_air2014}
 and human computer interactions~\cite{jaimes_cviu2007}. Vision-based HPE has been actively researched over the years and promising results have been reported  
on various challenging datasets recorded in common indoor and outdoor scenes \cite{sapp_cvpr2011,yang_pami2013,tokola_iccv2013,andriluka_cvpr2014,yang_cvpr2016,insafutdinov_eccv2016}. 
 
Even though state-of-the-art models such as  \cite{insafutdinov_eccv2016,yang_pami2013}  achieve impressive results on standard computer vision datasets, our experiments show that they do not necessarily generalize well to special environments like operating rooms (ORs). The quantitative results presented in Section \ref{sec:expRes} on data recorded during real surgeries show that there is still a large margin for improvement. These results are also in agreement with \cite{kadkhoda_media2016}, who reported that the Kinect skeleton tracker \cite{Shotton_PAMI2012}, which has been successfully used in the game industry, does not generalize well to the OR environment. The main issues are incorrect background subtraction and the mixing of the body parts belonging to different persons. Our supplementary video presents qualitatively how the aforementioned approaches perform on our OR data.

\begin{figure}[tb]
\centering
\setlength{\tabcolsep}{2pt}
\renewcommand{\arraystretch}{1}
\begin{tabular}{c c c}
\includegraphics[width=0.32\columnwidth]{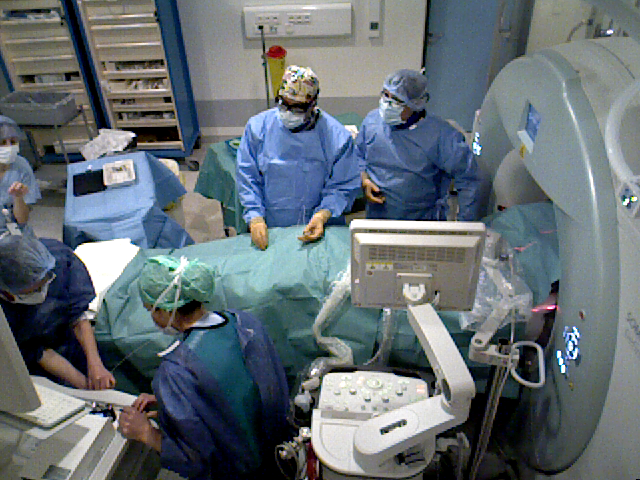} & 
\includegraphics[width=0.32\columnwidth]{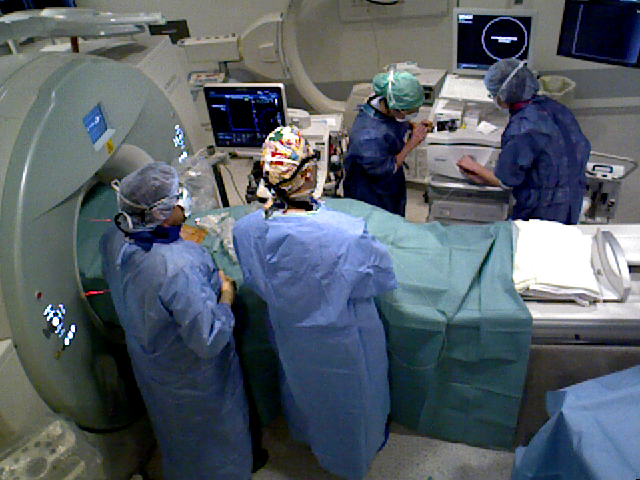} & 
\includegraphics[width=0.32\columnwidth]{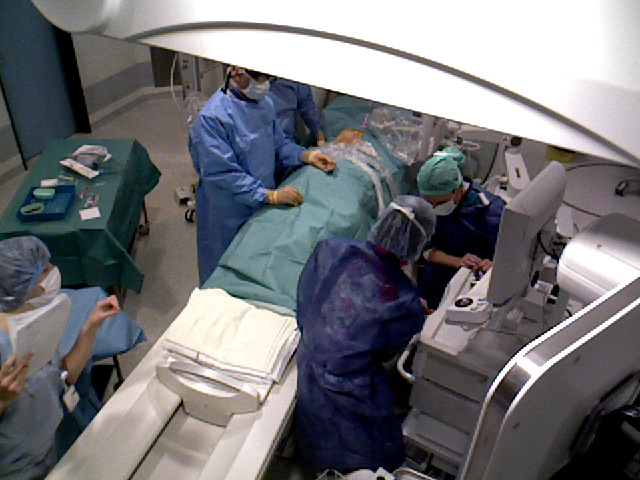} \\
\includegraphics[width=0.32\columnwidth]{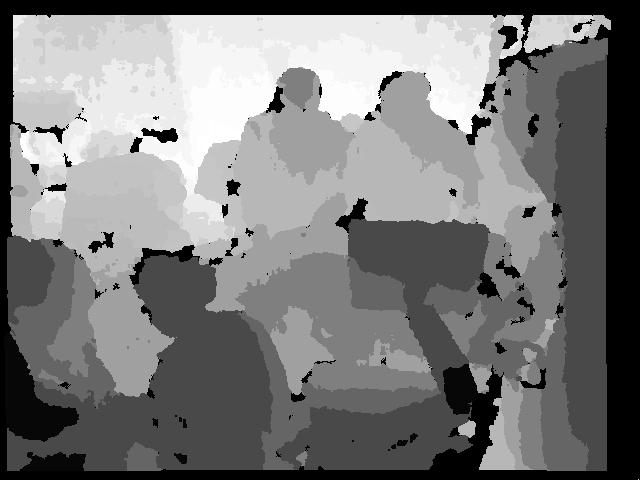} &
\includegraphics[width=0.32\columnwidth]{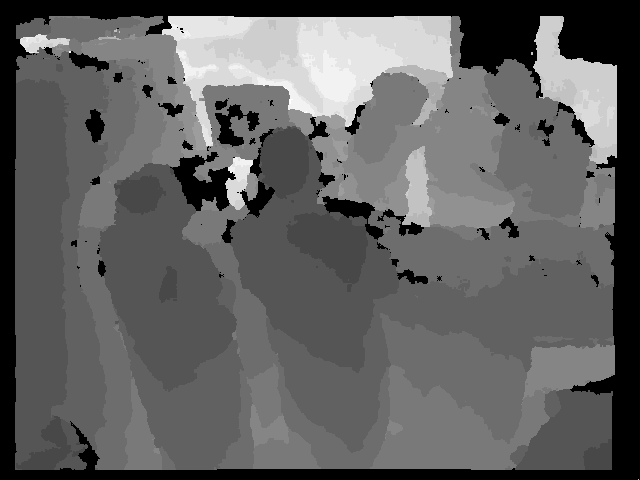} &
\includegraphics[width=0.32\columnwidth]{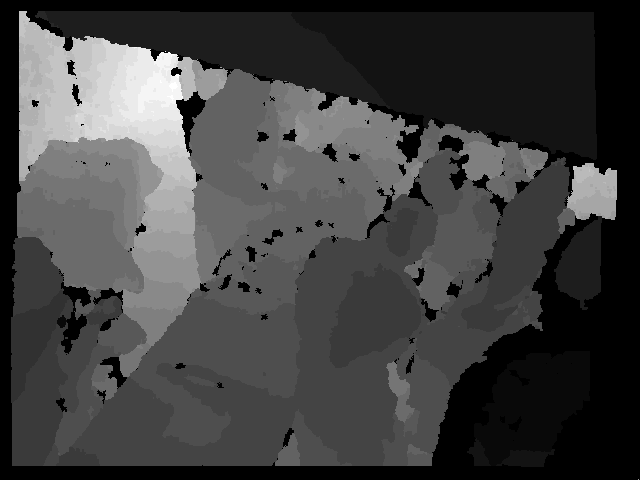}\\
\end{tabular}
\caption{Synchronized pairs of color and depth images from a novel multi-view operating room dataset. The images are recorded during live surgeries using a three-view RGB-D camera system.}
\label{fig:rgbdFrame}
\end{figure}

We believe that the drop in performance is caused by the inherent visual challenges present in such an environment. Figure~\ref{fig:rgbdFrame} shows an operating room during real surgeries: one can notice that the visual appearances of many different surfaces are very similar, which makes it difficult to distinguish persons from the background. People are also wearing loose and textureless gowns that make it hard to discriminate body parts. In addition, the camera positioning possibilities are very limited due to the objects and equipment that need to be frequently displaced in the room, such as ceiling mounted articulated arms, screens and the respiratory tower. 
Furthermore, performing a surgery requires the collaboration of multiple people, which increases the risk of occlusions.

In \cite{kadkhoda_media2016}, we have proposed an approach based on pictorial structures for clinician detection and pose estimation. The approach relies on a pair of color and depth images captured from a single viewpoint using a low-cost RGB-D sensor similar to the Microsoft Kinect camera~\cite{Shotton_PAMI2012}. That work has demonstrated that the combination of color and depth information along with the use of depth information to model 3D constraints between neighboring body parts greatly improves the pose estimation results. Following the findings of \cite{kadkhoda_media2016}, we propose in this paper to use a multi-view system based on RGB-D cameras. The system uses three cameras in order to capture the environment from complementary views and to reduce the risk of occlusions in such a cluttered environment. We show that the advantages of using depth maps in such a multi-view approach go beyond the mere generation of improved appearance features.

In a multi-view RGB system, correspondences across views are traditionally established by relying on appearance similarity and triangulation \cite{belagiannis_eccv2014,gall_ijcv2010,
amin_pr2014,belagiannis_mva2016}, which is unreliable in OR environments containing many surfaces that are visually similar. Instead, the depth data enables us to back-project points to 3D and is not affected by the visual appearance of the surfaces in the scene \cite{kadkhoda_media2016}. It also enables us to back-project points that are only visible in one view, while in multi-view RGB systems, points should be visible in at least two views. 

Current multi-view human pose estimation approaches have been proposed either for single-person scenarios  \cite{gall_ijcv2010,burenius_CVPR2013,hofmann_ijcv2011,amin_pr2014} or for multi-person scenarios in which the number of persons is known in advance \cite{luo_icpr2010,belagiannis_eccv2014,belagiannis_mva2016}. The approach proposed in this work makes no assumption about the number of persons in the scene. To this end, our approach first processes each view separately to detect putative skeletons. Next, {\it a priori} information about the environment, modeled using random forests, is applied to filter spurious skeletons. The resulting skeletons are then merged across views\footnote{We assume that the extrinsic parameters of the cameras are known.}. Finally, a novel energy function is optimized to incorporate evidence across views and update initial part positions directly in 3D.

Our single-view RGB-D pose estimation approach extends 3D Pictorial Structures (PS) \cite{kadkhoda_media2016} by incorporating Convolutional Neural Networks (ConvNets) for the part detection \cite{insafutdinov_eccv2016}. ConvNets have recently enjoyed a great success in solving many vision-based tasks including human pose estimation \cite{toshev_CVPR2014,tompson_cvpr2015,schmidhuber_nn2015,insafutdinov_eccv2016,pishchulin-cvpr2016}. They are capable of learning strong detectors that can incorporate a wide image context through deep network architectures with large receptive fields \cite{wei_cvpr2016}. Mutual spatial constraints among body parts are however not explicitly modeled, even though they are essential to guarantee joint consistency in the predicted body configuration, especially in multi-person and cluttered environments such as ORs. 
Therefore, we use a deep ConvNet-based part detector constructed for RGB-D data in conjunction with a 3D pairwise dependency model to enforce body kinematic constraints directly in 3D. This is in contrast to current methods that rely on 2D displacement \cite{yang_cvpr2016,jain_iclr2014,tompson_nips2014} or visual similarities \cite{insafutdinov_eccv2016} among body joints. Enforcing body kinematic constraints in 3D is crucial to reliably estimate body part configurations of different individuals who are close to each other in the projected 2D image and are visually similar.

Incorrect detections and occlusions can however result in spurious skeleton candidates in each view that can mislead the multi-view merging algorithm. We argue that in a specific environment like the operating room, {\it a priori} information about the room should be leveraged to identify spurious candidates. Therefore, we also propose a method to learn a prior on the 3D body kinematic and room layout constraints. This prior, based on random forest, is used to recognize and remove skeletons with unlikely 3D shapes or positions. Relying directly on high level 3D skeleton information enables the model to better explore the {\it a priori} information of the OR and to build a stronger prior compared to traditional pose priors that are based on the displacement or visual similarity among parts \cite{yang_cvpr2016,insafutdinov_eccv2016,kadkhoda_media2016}. 

This paper investigates multi-person multi-view pose estimation using RGB-D data and makes the following contributions. First, we extend the 3D pictorial structures of \cite{kadkhoda_media2016} to use a ConvNet body part detector on RGB-D images. Second, we propose a random forest based method to automatically learn a prior to incorporate {\it a priori}  information about the environment. Third, we introduce a novel multi-view energy formulation to estimate 3D body configurations by leveraging depth data and reasoning across all views. Finally, we have evaluated the approach on a novel multi-view OR dataset generated from several days of recordings during {\it live surgeries}. 

\noindent\textbf{Related work.} 
Approaches using RGB-D data for human pose estimation are mostly based on background subtraction \cite{baak_iccv2011,Shotton_PAMI2012}, which is a very challenging task in the OR. An exception is \cite{buys_vcir2013}, which relies on random forests and color to cluster image pixels into body parts and then estimate body poses. However, the performance of the system degrades dramatically due to occlusions and clutter, which mislead the pixel classifier and the color-based clustering algorithm. The multi-view RGB-D systems that have been proposed also rely on background subtraction and usually address simple scenarios in laboratory setups \cite{xu_icip2016}.
For this reason, we focus in the following description on part-based approaches (using RGB images) that do not rely on background subtraction, since they are at the core of our method. Such approaches represent the human body as a set of body parts and model body kinematic constraints using a deformation model. The pictorial structures framework is the main part-based approach and has driven much of the progress in the field since its introduction in \cite{fischler_tc1997}. 

In single-view pictorial structures, exact inference was made tractable by the seminal work of \cite{felzenszwalb_ijcv2005}, which has been extended in different ways by either constructing a stronger body part detector \cite{eichner_bmvc2009,yang_pami2013} or by improving the deformation model \cite{sapp_cvpr2011,kadkhoda_media2016}. Most recently, with the availability of large training sets and high computational power, state-of-the-art results are obtained by using deep convolutional neural networks as body part detectors \cite{tompson_cvpr2015,insafutdinov_eccv2016,newell_eccv2016}. 

Single-view multiple human pose estimation is often performed in two steps: person detection followed by pose estimation \cite{gkioxari_cvpr2014,sun_iccv2011}. But, when people are in close proximity to each other, \eg in the operating room, a body part can be assigned to more than one person due to weak body part detections or occlusions. \cite{insafutdinov_eccv2016}, which is built on \cite{pishchulin-cvpr2016}, has proposed a multi-person pose estimation approach based on integer linear programing to jointly detect people and estimate their body part configurations. In this approach, part detection is performed using deep ConvNets and interpart pairwise constraints are enforced based on 2D displacement and appearance similarity between body parts. But, image-based pairwise constraints are not very discriminative in operating rooms since people are wearing textureless gowns with similar colors (see Figure \ref{fig:rgbdFrame}). Instead, \cite{kadkhoda_media2016} uses more discriminative 3D pairwise constraints. We follow a similar formulation to \cite{kadkhoda_media2016}  using 3D pairwise constraints, but include a deep ConvNet-based detector in our model instead of a support vector machine with engineered visual features. 

Most work on multi-view human pose estimation focuses on single-person scenarios in controlled environments to reduce the ambiguity of data association by relying on background subtraction or exemplar-based approaches \cite{gall_ijcv2010,hofmann_ijcv2011,sigal_ijcv2009}. In order to be robust to cluttered background, part-based approaches are used to generate part hypotheses per view and then estimate the body configuration in 2D \cite{amin_bmvc2013,amin_pr2014} or in 3D \cite{burenius_CVPR2013}.

Recently, a PS approach has been proposed to estimate the poses of multiple persons in a multi-view setup \cite{belagiannis_CVPR2014}. The body pose estimation is performed in 3D by relying on 2D view appearance cues and  on multi-view cues computed through triangulation. This approach has been extended in \cite{belagiannis_eccv2014} to also include temporal cues. However, all these methods require prior knowledge of the number of persons present and have been evaluated on scenarios recorded in controlled laboratory environments that include people in upright poses only. In contrast, our method detects the number of persons and does not rely on triangulation that might not always be reliable in complex and cluttered environments. Moreover, we have evaluated our approach on a dataset recorded during real surgeries, which contains many visual challenges and where people exhibit a much wider range of articulations compared to those used in the aforementioned works.

A multi-view clinician pose estimation approach has been proposed in \cite{belagiannis_mva2016}. Background subtraction and tracking over the entire sequence are used to find the trajectories of the persons and to localize them using bounding boxes. Then, the approach presented in \cite{belagiannis_CVPR2014} and a ConvNet-based RGB part detector are used to estimate the pose of each individual given the bounding boxes. To evaluate this approach, two sequences containing a constant number of individuals have been recorded during two medical procedures simulated by actors. In contrast, our approach relies on a multi-view set of images from a single time-step. Moreover, our dataset has been generated from four days of recordings during live surgeries, which is more challenging due to: (1) a high variation in number of persons per frame and (2) the presence of many movable objects in the scene, which makes it difficult to compute the foreground.

\section{Method}

We start this section by recapitulating the clinician pose estimator of \cite{kadkhoda_media2016} and then present the different components that lead to our multi-view RGB-D approach.


\subsection{Single-view body pose estimator}
\label{sec:sv_bpe}
In \cite{kadkhoda_media2016}, we have presented a pictorial structures model to estimate body configurations on RGB-D images. This model represents the body as a set of $n$ joints and learns multiple mixtures of parts to be robust to appearance changes. The model uses ten body joints to indicate upper-body poses, since lower body parts are often occluded in operating rooms. A body configuration is specified by a pair $(l,t)$, where $l=\left\lbrace l_1 ... l_n  \right\rbrace$ indicates the 2D positions of the body joints and $t_i$ belongs to a set of $m$ possible mixture types $t=\left\lbrace t_1 ... t_n  \right\rbrace$ for each body joint. The pose estimation is defined as an energy minimization over a tree-structured graph $G=(V,E)$, whose nodes are the body joints and whose edges indicate dependencies between joints. The body joint dependencies are defined following the human body skeleton. Given a pair of aligned color and depth images denoted by $I$ and $D$, respectively, the score associated with a body configuration $(l,t)$ is defined as: 
\begin{dmath}
\label{eq:Energy3D}
S(I,D,l,t) = \sum_{i\in V}\phi(I,D,l_i) + \sum_{ij\in E} w_{ij}^{t_i,t_j}.\psi(D,l_i,l_j)  + \sum_{i\in V} b_i^{t_i} + \sum_{ij\in E} b_{ij}^{t_i,t_j},
\end{dmath}
where the first term is the appearance model, which is also referred to as the part detector, and the second term is the deformation model that enforces pairwise dependencies between body joints. The last two terms are part type compatibility score functions, where $b_i^{t_i}$ captures the score of assigning a particular mixture type to part $i$ and $b_{ij}^{t_i,t_j}$ is the score associated with the co-occurrence of a particular pair of part types. The compatibility score functions and $w_{ij}^{t_i,t_j}$ are the model parameters. These parameters are learned using a structured support vector machine formulation.   

The part detector assigns a confidence score for placing the body joint $i$ at image location $l_i$. \cite{kadkhoda_media2016} relies on handcrafted features, namely  Histogram of Oriented Gradients (HOG) and Histogram of Depth Differences (HDD). In this work, we compute the part detection scores using the Deep ConvNet model presented in Section \ref{sec:ConvNet}.  

The deformation model is parametrized by  $w_{ij}$ and $\psi(D,l_i,l_j)$. The weights $w_{ij}$ encode the deformations between pairs of joints. The relative displacement of joint $i$ w.r.t. joint $j$ is captured by $\psi(D,l_i,l_j) = [|d_{3D}|, {dc},{{dc}}^2,{dr}, {{dr}}^2]^T$, where $|d_{3D}|$ is the absolute 3D Euclidean distance between the joints and $(dc,dr)$ are the relative displacements along columns and rows of the image. The 3D joint positions are computed by back-projecting 2D joints into 3D using the depth image. This term enables the model to incorporate more reliable 3D body part lengths for assembling body joints, which is important in order to discriminate between detections on the surface of a person and detections on the background. 

Estimating the body configuration in this model corresponds to finding the optimal body joint positions and mixture types given color and depth images. This process, called inference, computes $(l^*,t^*) =\argmax_{l,t} S(I,D,l,t)$. It is performed in 3D using an efficient algorithm that makes exact inference tractable. For more information, we refer the reader to \cite{kadkhoda_media2016}.

\subsubsection{ConvNet-based RGB-D body part detector}
\label{sec:ConvNet} 
Motivated by the great success of deep convolutional neural networks in recent years \cite{schmidhuber_nn2015,he_arXiv2015,tompson_cvpr2015,insafutdinov_eccv2016,newell_eccv2016}, we propose to use RGB-D body part detectors based on deep ConvNets in order to automatically learn feature representations instead of relying on engineered feature representations such as HOG or HDD. To this end, we build on the very deep residual network \cite{he_arXiv2015}, which has recently been used for part detection and shown promising results \cite{insafutdinov_eccv2016}. The body part detection is formulated as a multi-label classification problem, where a set of $n$  scores is generated at each image location to denote the probability of part presence. The scores are obtained by using sigmoid activation functions on the output neurons. We adapt the network to learn body part detectors for pairs of color and depth images. We change the input layer to accept four dimensional data (\ie three color channels and depth channel). We also change the $\mathtt{res3d}\_\mathtt{pose}$ layer to generate part score maps for ten upper-body parts instead of the fourteen full body parts. During pose estimation, we use the ConvNet-based body part detector to predict confidence scores for all parts at every image locations. Hereafter, we refer to this HPE model as {\it Deep3DPS}.

\noindent\textbf{Fine-tuning.} We initialize the network from the pre-trained model of \cite{insafutdinov_eccv2016}, which is trained on the {\it MPII Human Pose} dataset. We fine-tune the network on the single-view clinician pose dataset from \cite{kadkhoda_media2016} that consists of 1451 RGB-D frames including 1991 persons using the Caffe framework \cite{caffe}. We scale the images down to 85\% and use a batch size of two. Similarly to  \cite{insafutdinov_eccv2016}, we generate target training score maps for all body joints by assigning the positive label 1 for all image locations within 15 pixels to the ground truth location and negative label 0 otherwise. During training, we use all positive samples and keep at most three times more negative samples. The network is trained with  cross entropy loss and stochastic gradient descent for $50$k iterations. The initial learning rate is set to $5\times10^{-5}$ for the adapted layers and $5\times10^{-6}$  for the rest. This yields the best results in our experiments. In \cite{insafutdinov_eccv2016}, the network is trained for three tasks: body part detection, location refinement, which is the relative row and column displacement from a scoremap location to the ground truth, and regression to other parts. However, training for the last two tasks did not yield any performance improvement during our experiments. We therefore only train for the body part detection task.

\subsection{Random forests based prior}
\label{met:prior}

To design a robust method, we believe that it is essential to include priors specific to the environment. Even though a general body kinematic prior is included in the pose estimation model through pairwise constraints, it cannot be guaranteed that these constraints are always properly enforced due to the high complexity of the pose estimation model that predicts human poses directly from image pixel values. In addition, this prior only captures body kinematic constraints and does not incorporate {\it a priori} information about the environment. In an environment like the OR, constraints such as possible human poses and possible locations can also be used to improve the reliability of the method. Such constraints cannot be easily handcrafted. Furthermore, including them in the pose estimation model would need higher-order dependency terms. Adding such terms would increase the number of model parameters and, more importantly, dramatically increase the complexity of the inference algorithm.
We therefore propose to automatically learn the prior, which we formulate as a binary classification problem that takes a skeleton estimated by the single-view detector as input and outputs whether this skeleton corresponds to a spurious detection or not.


We base our approach on Random Forests (RF), which are an ensemble of decision trees consisting of two types of nodes: split and leaf nodes. In each split node, a decision function forwards samples to one of the branches until they finally reach a leaf node containing a prediction function. In our case, we use RF with binary trees and the mean over all predictions to aggregate the votes across all trees. 
The trees are learned automatically given a labeled training set, which we construct using the skeletons estimated by our single-view pose estimator on a set of images for which ground-truth is available.
The detected skeletons are compared to the ground truth using the probability of correct keypoints (PCK) metric, which is commonly used for evaluation in multiple-person pose estimation \cite{yang_pami2013,kadkhoda_media2016,pishchulin-cvpr2016}. We label a detected skeleton as positive if the head, neck, and left and right shoulders are correctly localized according to PCK. 

For RF training, we propose to combine various features computed from the 3D skeletons, which are all expressed in the common room reference frame. The reference coordinate system is chosen w.r.t. the operating table in default position, which makes the prior generalizable to other ORs. This enables our prior to encode two types of information: room layout and possible clinician poses.
Certain parts of the room, such as the floor or the ceiling, are for instance not expected to have clinicians or certain body parts. Thus, as first set of features, we use the positions of the 3D body parts to enable the RF to build an internal representation of their spatial occupancy probability. To capture the set of possible human poses in the OR, we include a second set of features, namely the relative 3D displacements between all pairs of body joints. The prior also serves to verify 3D part lengths and exclude incorrect skeletons that may occur due to weak detections and foreground/background confusions. As third feature, we include the detection score of the individual skeleton to incorporate detection confidence. To enable our prior to better encode high-level information, we use the RF method in a multi-layer scheme, referred to as auto-context in the machine learning literature \cite{tu_pami2010}. A multi-layer model is learned, where the first RF layer is constructed using only the three aforementioned types of features, while the other layers use another extra feature that is the classification confidence generated by the previous RF layer. 


\subsection{Multi-view human pose estimation}


\subsubsection{Multi-view fusion}
\label{sec:fusion}
The objective of the multi-view fusion is to combine the 3D skeletons across all views. For a given {\it frame}, defined as a set of RGB-D images recorded from all cameras at the same time step, detections from all views are first put in a set. The two closest skeletons that do not originate from the same view are then merged. This procedure is iterated until no pair of merging candidates is left in the set, where the condition for merging two skeletons is that the distance between their heads and the distance between their necks are both smaller than a constant $T_s$. Since the left/right side labels of the individual detections are not always reliable, to ensure a consistent merging of the 3D joints we use the 3D positions of the shoulders to find the correct association between the two skeletons. Finally, for all skeletons resulting from a merging step, the left and right side labels are set based on a majority vote among the supporting skeletons. If a merged skeleton originates from only two supporting skeletons, which do not agree on the side label, we set the side according to the skeleton with highest confidence. 

As a result, we obtain a set of initial 3D skeletons generated from skeletons coming from one or more views. Then, a new multi-view energy function, presented next, is used to drive the body parts towards their optimal 3D locations by jointly optimizing over all views. 


\subsubsection{Multi-view RGB-D Optimization}

We formulate our multi-view RGB-D approach as an energy minimization over the same graph $G$ as in Section~\ref{sec:sv_bpe} and define the energy function $E(\Delta)$ over the graph as:
\begin{dmath}
\label{eq:MRF_multiView}
E(\Delta) = \sum_{i \in V} \Big( \lambda_1 .\Phi^{conf}(\delta_i) + 
\lambda_2 .\Phi^{depth}(\delta_i) \Big) + 
\sum_{(i,j) \in E} \Psi_{i,j} (\delta_i, \delta_j), 
\end{dmath}
where $\lambda_1$ and $\lambda_2$ are weighting coefficients, $\Delta = \lbrace \delta_1 ... \delta_n  \rbrace$ is a set of displacement labels for all body parts, $\delta_i \in \mathord{\mathbb R}^3 $ is a 3D displacement offset for part $i$, $\Phi(.)$ are the unary potentials and $\Psi_{i,j} (\delta_i, \delta_j)$ is a pairwise dependency term enforcing body physical constraints. 

The first term in \eqref{eq:MRF_multiView} incorporates part detection confidence scores computed by the ConvNet part detector. Given the list of all views $views$, we define:
\begin{dmath}
\Phi^{conf}(\delta_i) = \sum_{\mathtt{v} \in views} conf\Big(proj\big(P(\delta_i), \mathtt{v}\big) \Big),
\end{dmath}
where $P(\delta_i)$ is the 3D position of part $i$ displaced by an offset $\delta_i$ and $proj(p_{3D}, \mathtt{v})$ projects the 3D point $p_{3D}$. In order to provide a smooth cost function, we compute the distance transforms of the deep ConvNet score maps using the generalized distance transform algorithm \cite{felzenszwalb_ijcv2005}. We find that this transformation is necessary to avoid local minima. $conf(p_{2D}) \in [0..1]$ is the value of the distance transform of the score map of part $i$ at location $p_{2D}$.
The second term is defined as: 
\begin{dmath}
\Phi^{depth}(\delta_i) = \sum_{\mathtt{v} \in views} 
\Big| D\Big(proj(P(\delta_i), \mathtt{v})\Big) - Z(P(\delta_i),\mathtt{v})\Big|,
\end{dmath}
where $D(p_{2D})$ is the depth value at image location $p_{2D}$, $Z(p_{3D}, \mathtt{v})$ is the z value of the 3D point $p_{3D}$ in the coordinate system of the view and $|.|$ is the absolute value operator. To reduce the effect of the noise present in the depth image, we smooth the depth image with a median filter of size $7{\times}7$px. This term quantifies the distance between the displaced 3D joint and the surfaces captured by the depth cameras.  Therefore, it can help to avoid placing parts in ghost 3D locations that do not correspond to any surface in the scene.
These two unary terms incorporate multi-view cues, where the RGB-D ConvNet is used to include image evidence and depth is used to integrate a reprojection cost across all views.

\begin{figure}[t]
\begin{center}
\setlength{\tabcolsep}{1pt} 
\begin{tabular}{ccccc}

\includegraphics[width=0.32\linewidth]{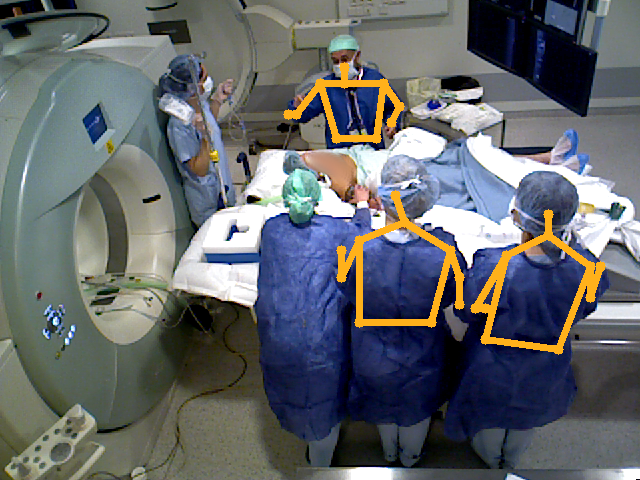} &
\includegraphics[width=0.32\linewidth]{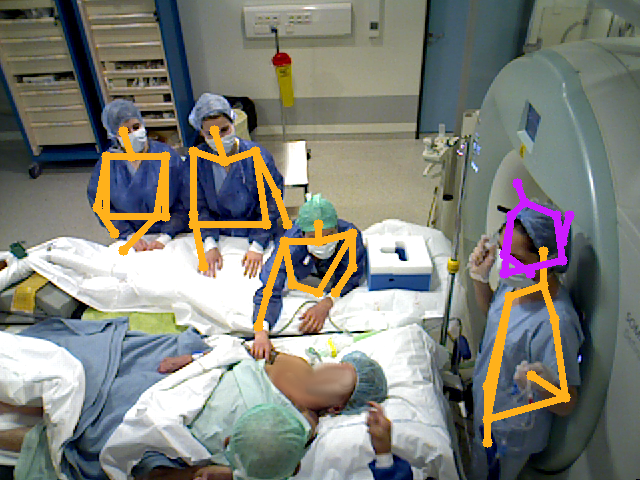} &
\includegraphics[width=0.32\linewidth]{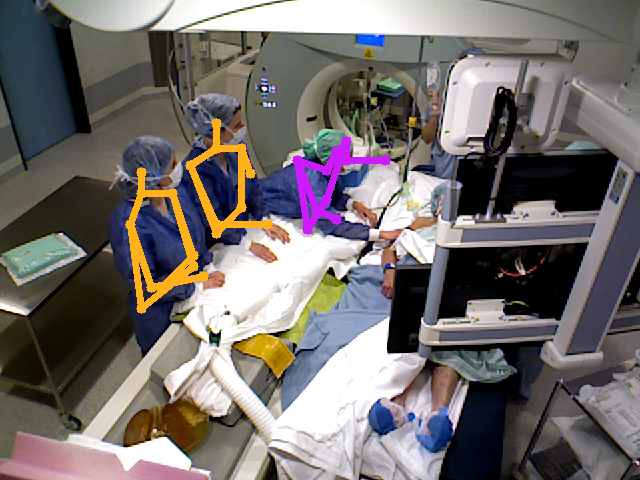} \\
\includegraphics[width=0.32\linewidth]{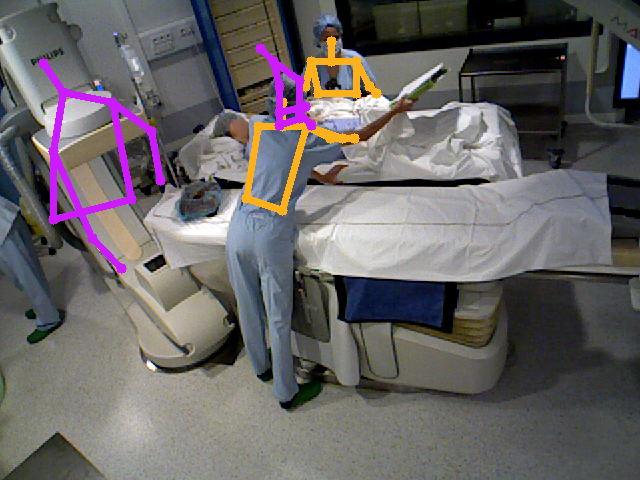} &
\includegraphics[width=0.32\linewidth]{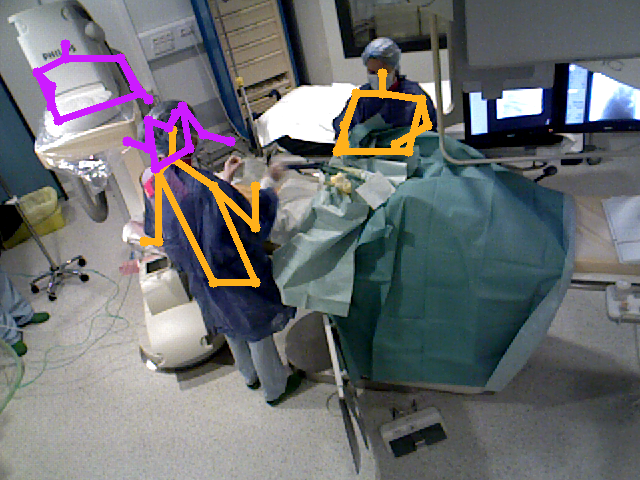} &
\includegraphics[width=0.32\linewidth]{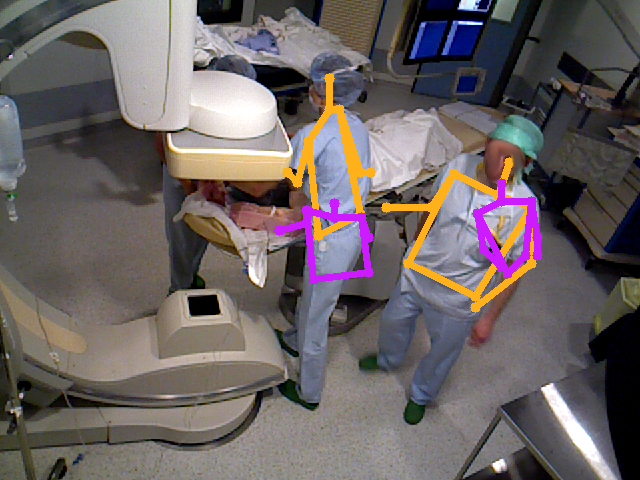} \\

\end{tabular}
\end{center}
   \caption{Multi-view examples illustrating the results of the RF-based prior. 
   Accepted skeletons are shown in orange and rejected skeletons in purple.} 
\label{fig:spuriousDet}
\end{figure}

The pairwise term is used to enforce kinematic constraints, namely body part lengths between pairs of joints. Let $\Psi_{i,j}$ be defined as:
\begin{dmath}
\Psi_{i,j} (\delta_i, \delta_j) = | \|P(\delta_i) - P(\delta_j)\| - \mu_{i,j}|,
\end{dmath}
where $\|.\|$ is the $\mathcal{L}_2$-norm and $\mu_{i,j}$ is the average distance between joints $i$ and $j$, \ie average part length. The average part lengths are computed over the entire training dataset. Note that since the body part lengths are relatively constant in 3D, it is here not needed to learn person-specific average part lengths.     

\noindent\textbf{Inference.} In order to recover 3D body part configurations, we need to perform inference in 3D. This problem corresponds to optimizing the energy function in Eq. \eqref{eq:MRF_multiView}. Note that using the optimization algorithm of \cite{kadkhoda_media2016} would require to construct a 3D state space that includes all 2D positions back-projected to 3D (amounting to the number of views multiplied by the size of the images) augmented with extra nodes for each back-projected node to account for occlusions. Such a large state space would degenerate the performance and slow down the inference. Similarly, the inference approach from \cite{burenius_CVPR2013} would limit us to use simple binary pairwise terms. Instead, we perform discrete optimization using the {\it fast-PD} algorithm \cite{komodakis_cviu2008}, which casts the optimization problem in an integer programming framework and exploits solutions form both primal and dual problems for efficiency. To perform the optimization, we define a set of discrete displacement labels $\mathcal{L}$ for each body joint by sampling densely from a cube centered at the initial joint position. The sampling function is parametrized by $(k,s)$, where $k$ is the number of samples along each 3D direction and $s$ is the step size between the samples. 
We perform the optimization iteratively by starting with a coarse label set that covers a large 3D space. At the end of each iteration, we update the part positions based on the displacement labels and then generate a finer label set for the next iteration.

\section{Experimental results}
\label{sec:expRes}
\noindent\textbf{Datasets.} We have generated a novel multi-view RGB-D dataset, illustrated in Figure~\ref{fig:rgbdFrame}, by recording all activities in an operating room for four days.
For quantitative analysis, the 3D upper body poses of 1378 clinicians have been manually annotated in 741 multi-view frames that are evenly distributed across the dataset. 
All clinicians who have more than $50\%$ of their upper-body parts visible in at least one view have been annotated in these frames.
The annotations are performed using a tool that displays a 3D point cloud reconstructed from all three views as well as the corresponding individual 2D images. This tool allows the user to move the body joints either in the 2D views or in the 3D point cloud. Whenever a joint is moved in 2D, the 3D position of the corresponding joint in the average 3D skeleton is updated using the depth map and then reprojected back to all views. Thus, the annotator can verify the correctness of the annotated skeletons using both 3D visualization and 2D reprojections across the views.

In order to have a fair comparison with our method in  \cite{kadkhoda_media2016}, the single-view dataset of \cite{kadkhoda_media2016} is used for training all 3DPS single-view pose estimation models and for fine-tuning the network. For all models, evaluation is performed on the new multi-view dataset.  
As the multi-view dataset is used for random forest training, to evaluate the model, a 4-fold leave-one-out cross-validation is performed, where one folds is used for testing and the rest for training. The evaluation reports the average results of the cross-validation.

\begin{table}[t]
\small
\centering
\setlength{\tabcolsep}{2pt} 
\renewcommand{\arraystretch}{0.95}
\begin{tabular}{l c c c c c |c}
	\toprule
	Setting  & Head  & Shld & Elbow & Wrist & Hip & Avg \\
	\midrule
Deep3DPS (DeeperNet)  & 89.6 & 56.5 & 50.6 & 54.3 & 42.9 & 58.8 \\
Deep3DPS (RGB) & {\bf 93.7} & 74.9 & 69.6 & 71.8 & 66.6 & 75.3 \\
Deep3DPS (Depth) & 91.0 & 75.0 & 69.1 & 68.0 & 63.2 & 73.2\\
Deep3DPS (RGBD)& 93.4 & {\bf 77.0} & {\bf 71.5} & {\bf 73.7} & {\bf 69.1} & {\bf 76.9}\\
\hspace{3em}+Auxiliary tasks & 91.4 & 72.1 & 64.9 & 68.4 & 63.5 & 72.1\\
\midrule
3DPS (IHOG+HDD) \cite{kadkhoda_media2016} & 90.8 & 74.2 & 62.2 & 63.4 & 57.5 & 69.6 \\
\midrule
Insafutdinov et al.	\cite{insafutdinov_eccv2016}\footnote{We use the model that was made publicly available by the authors.} & 91.1 & 53.7 & 47.5 & 50.1 & 38.4 & 56.2 \\
Yang and Ramanan \cite{yang_pami2013}\footnote{The model is trained on the Buffy dataset \cite{eichner_ijcv2012} using the public implementation of the approach.} & 30.4 & 35.2 & 19.6 & 24.3 & 16.7 & 25.2\\
%
			\bottomrule
	\end{tabular}
	\vspace{0.2cm}
	\caption{Pose estimation results of several single-view approaches using PCK metric. }
	\label{tab:pckResMVDS}
\end{table}

\subsection{Single-view pose estimation}

\begin{figure}[t]
\begin{center}
\setlength{\tabcolsep}{1pt} 
\begin{tabular}{ccccc}

\includegraphics[width=0.32\linewidth]{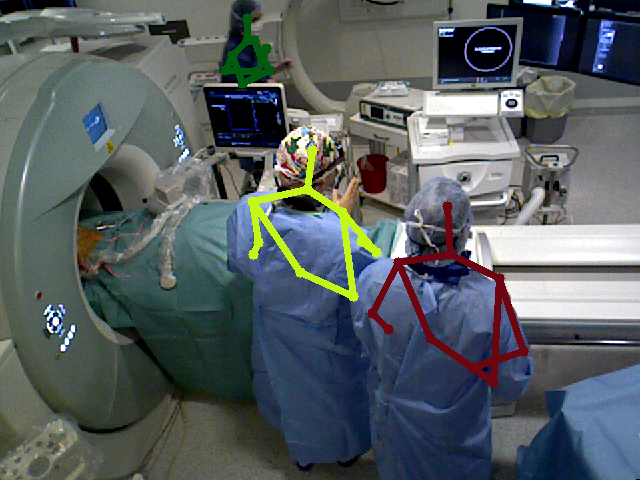} &
\includegraphics[width=0.32\linewidth]{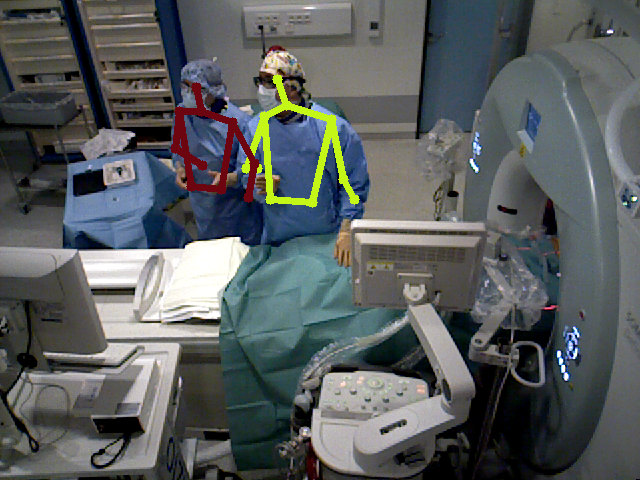} &
\includegraphics[width=0.32\linewidth]{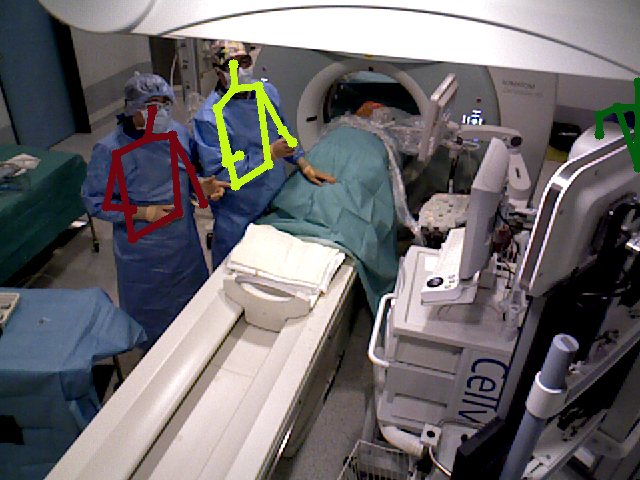} \\
\includegraphics[width=0.32\linewidth]{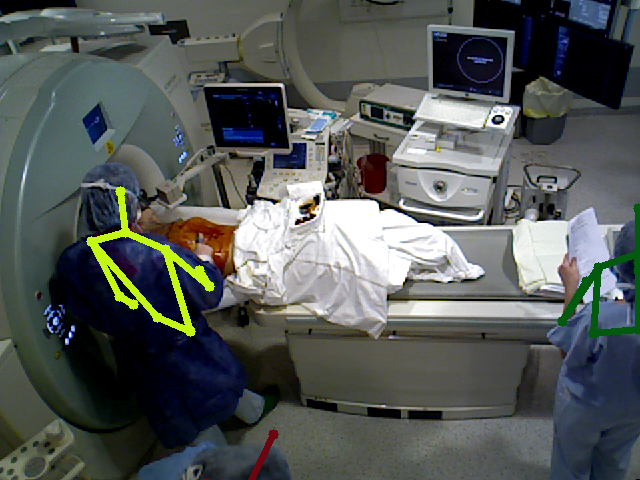} &
\includegraphics[width=0.32\linewidth]{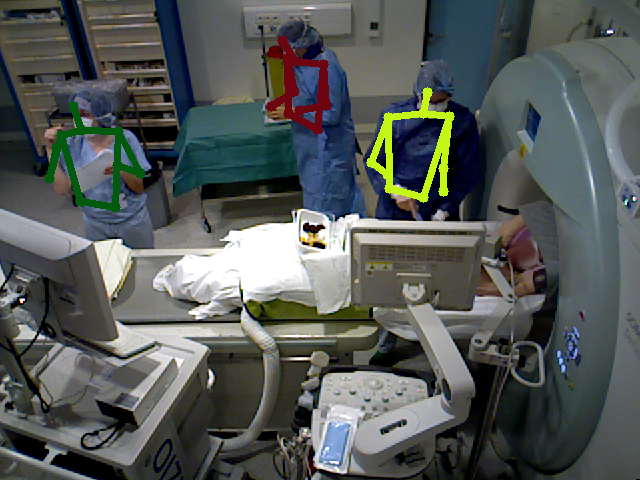} &
\includegraphics[width=0.32\linewidth]{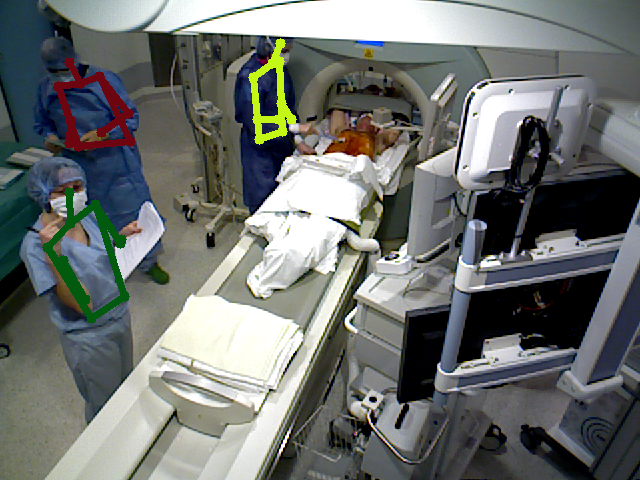} \\

\end{tabular}
\end{center}
   \caption{Examples of multi-view pose estimation results. Each row shows a multi-view frame. The 3D skeletons obtained after multi-view energy optimization are projected to the views.}   
\label{fig:qualitativeRes}
\end{figure}

Table \ref{tab:pckResMVDS} reports the performance of different models on the multi-view dataset using the {\it PCK} metric \cite{yang_pami2013,kadkhoda_media2016,insafutdinov_eccv2016}. All  3DPS models have been trained on the single-view dataset used in \cite{kadkhoda_media2016}. The 3DPS method using the pre-trained network of \cite{insafutdinov_eccv2016} as body part detector, referred to as {\it DeeperNet}, achieves a better performance compared to the full {\it DeeperCut} approach from \cite{insafutdinov_eccv2016} that estimates the body poses via a joint optimization across all people. These results indicate that in an environment with many visually similar surfaces, a 3D deformation model, even with tree-structured graph, is more reliable than a fully connected deformation model which relies on appearance and 2D displacement constraints. Fine-tuning the network on the single-view dataset significantly improves the results ({\it Deep3DPS (RGB)}: 75.3\% vs. 58.8\% PCK), as it allows the network to adapt its representation for learning a better encoder for such an environment. We have also trained the network to detect body parts using only depth data, {\it Deep3DPS (Depth)},  which achieves competitive results. The best performance is obtained when the network relies on both color and depth images: the resulting model, called {\it Deep3DPS (RGB-D)}, is therefore used as single-view pose estimator during the rest of the experiments. But, we observe that on this data, training the network for the auxiliary tasks suggested in \cite{insafutdinov_eccv2016}, namely location refinement and regression to other parts, degrades the performance. We believe that this is due to both a much smaller training set and to the strong foreshortening of the body parts because of the top views of the cameras.  

As baseline, we evaluate the performance of the best 3DPS model from \cite{kadkhoda_media2016}, which relies on a 3D deformation model similar to our approach, but with handcrafted color and depth features.
Our best model improves the performance over this baseline by ${\sim}7\%$ on the same experimental setup. This highlights the benefits of deep ConvNets in constructing more discriminative body part detectors by automatically learning feature representations and also incorporating a wider context.
Evaluation of state-of-the-art RGB models \cite{insafutdinov_eccv2016,yang_pami2013} trained on common computer vision datasets shows that they do not generalize to the OR environment due to both loose clinical clothes and the presence of many visually similar surfaces.

\begin{table*}[tb]
\small
\setlength{\tabcolsep}{6pt} 
\renewcommand{\arraystretch}{1.1}
\centering
	\begin{tabular}{lccccccccc}
		\toprule
\multirow{2}{*}{\begin{tabular}[c]{@{}l@{}}Part\\ name\end{tabular}} & \multicolumn{3}{c}{One view} & \multicolumn{3}{c}{Two views} & \multicolumn{3}{c}{Three views} \\ 
\cmidrule(lr){2-4} \cmidrule(lr){5-7} \cmidrule(lr){8-10}
                  & initial  & after opt. & opt.-$\Phi^{depth}$  & initial  & after opt.  & opt.-$\Phi^{depth}$   & initial  & after opt. & opt.-$\Phi^{depth}$    \\ 	 \midrule
Head  &  $7\pm4$ & $7\pm4$ 	   & $7\pm4$   & $6\pm3$ & $6\pm3$ &    $6\pm3$ &  $5\pm2$ & $5\pm2$ & $5\pm2$ \\
Neck  &  $7\pm4$ & $7\pm4$     & $7\pm4$   & $5\pm3$ & $5\pm3$ &    $5\pm3$ &  $4\pm2$ & $4\pm2$ & $4\pm2$\\
Shld  &	 $25\pm25$ & $19\pm16$ & $21\pm18$ &  $22\pm16$ &$15\pm10$ &  $19\pm13$ &  $14\pm14$ & $10\pm7$ & $12\pm9$ \\
Hip	  &	 $28\pm22$ & $27\pm19$ & $28\pm20$ & $24\pm13$ & $23\pm13$ &  $24\pm14$ & $18\pm10$ & $17\pm9$ & $18\pm10$\\
Elbow &	 $31\pm22$ & $27\pm19$ & $30\pm21$ &  $30\pm18$ & $23\pm15$ & $27\pm18$ & $19\pm14$& $16\pm11$ & $18\pm14$\\
Wrist &	 $42\pm34$ & $32\pm21$ & $35\pm24$ &  $34\pm22$ & $25\pm16$ &  $28\pm18$ & $24\pm18$ & $18\pm13$ & $20\pm15$\\ 
\midrule
avg$\dagger$ &	 $32\pm26$ & $\mathbf{26\pm19}$ & $29\pm21$ &  $28\pm17$ & $\mathbf{22\pm14}$ &  $25\pm16$ & $19\pm14$ & $\mathbf{15\pm10}$ & $17\pm12$\\
\bottomrule
\end{tabular}
	\vspace{0.2cm}
	\caption{Mean and standard deviation of 3D part localization error in centimeter. The results are presented as a function of the number of supporting views used to generate the initial 3D skeletons (distribution: 1 view: 30\%; 2 views: 43\%; 3 views: 27\%). $\dagger$ The average is computed for all parts except the head and neck since they are not included in the optimization. See Section \ref{exp:multiview} for details.}
	\label{tab:localization}
\end{table*}

\subsection{Random forest based prior} 

\begin{figure}[t]
\begin{center}
\renewcommand{\arraystretch}{0.6}
\setlength{\tabcolsep}{1pt} 

\begin{tabular}{ccc}
\includegraphics[width=0.465\linewidth]{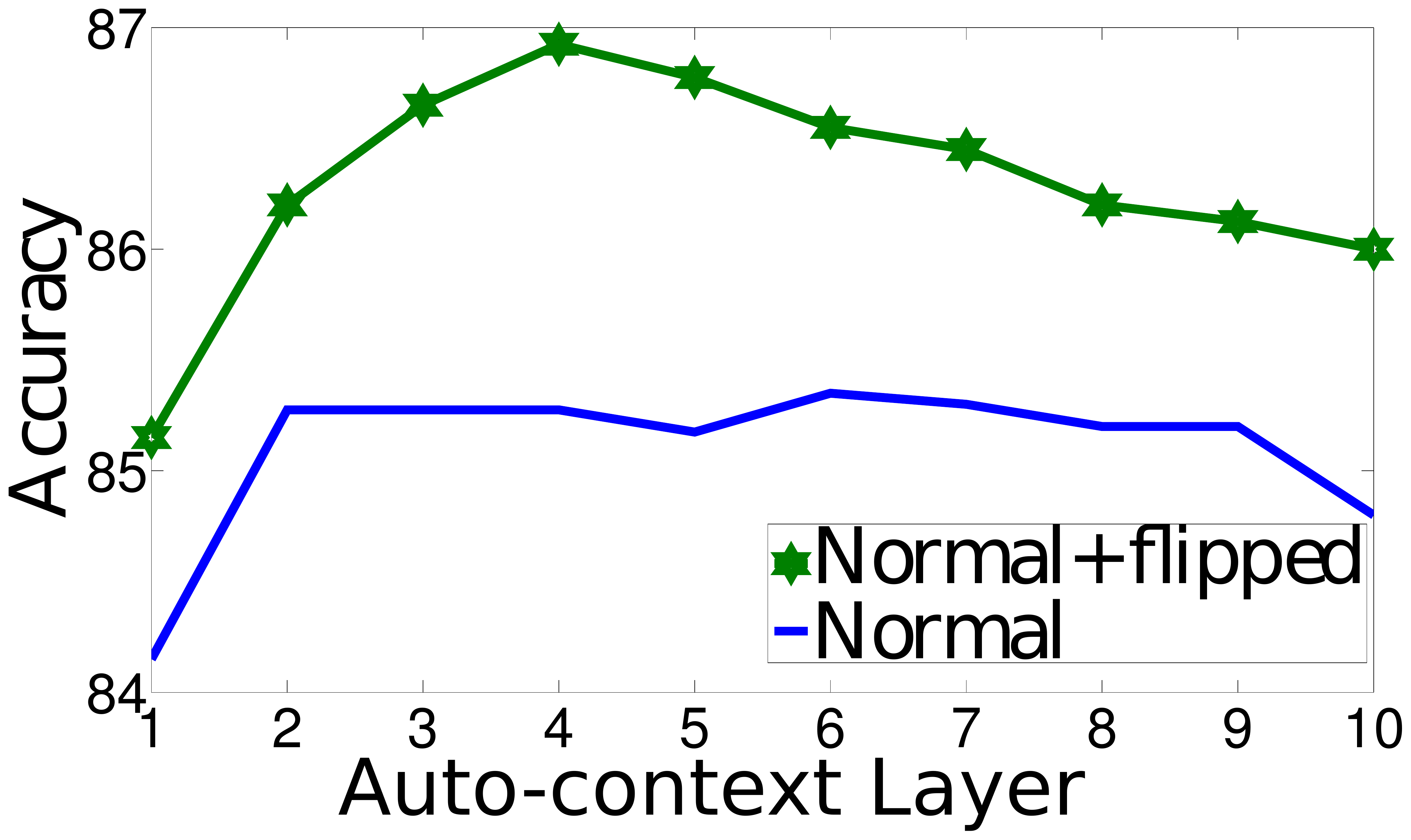} & &
\includegraphics[width=0.50\linewidth]{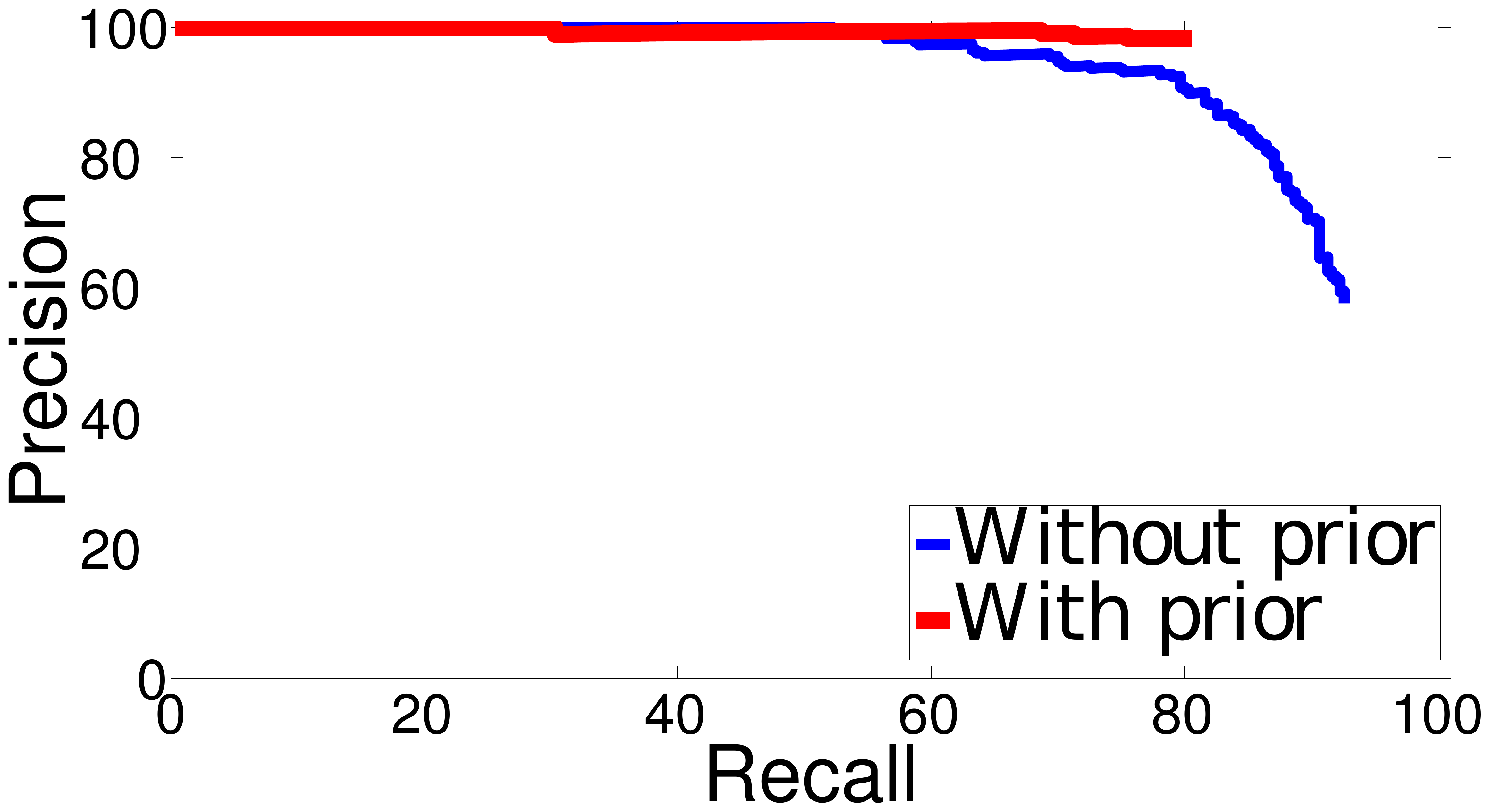} \\
\vspace{-0.2cm}
{\hspace{0.15cm} \scriptsize (a)} & & \hspace{0.25cm} {\scriptsize (b)}\\
\end{tabular}
\end{center}
\caption{(a) Accuracy of the RF-based prior in detecting spurious skeletons. (b) Precision-recall curves for 3D clinician detection.}
   
\label{fig:resPrior}
\end{figure}

The Deep3DPS (RGB-D) model 
 is applied to detect skeletons in each view of the multi-view dataset separately. The skeletons are back-projected into 3D and transformed into a common reference frame. We use these 3D skeletons to train our prior, as explained in section \ref{met:prior}. We also augment the data by flipping the skeletons to exchange the left-right body parts. Due to the small size of the training set, we learn 100 shallow trees with a maximum depth of 10.
Figure~\ref{fig:resPrior}(a) shows the detection accuracy of the RF-based method in distinguishing spurious detections. The results show that the method always detects valid skeletons with an accuracy superior to $84\%$. One observes that augmenting the training set by flipping the skeletons consistently improves the results by enabling the forest to learn a richer prior model that is not confused by the noisy side detections. The results also indicate that auto-context enhances the performance up to the fourth iteration and then tends to overfit. We therefore use the output of the RF trained on the augmented training set at the fourth layer to identify spurious skeletons during the remaining evaluations. Figure~\ref{fig:spuriousDet} illustrates the results of the approach on sample frames from the multi-view dataset and also on a few frames recorded in a different room from totally different viewpoints. It can be seen that the method has correctly identified spurious skeletons in both datasets and generalizes well. The proper generalization is due to the fact that the reference frame is defined on the floor at the center of the operating table, the main element in any operating room. 


%

\subsection{Multi-view 3D detection and pose estimation}
\label{exp:multiview}

We set $T_s$ to 30 cm to avoid merging skeletons across persons who are close to each other. We evaluate 3D clinician detection using the precision-recall curves. A detection is accepted as a true positive if the distance between the ground-truth and the detection is below 30 cm for both the head and neck. We use the fusion algorithm described in Section \ref{sec:fusion} to generate a set of 3D skeleton candidate per frame. Figure~\ref{fig:resPrior}(b) shows the 3D clinician detection results after multi-view fusion with and without the RF-based prior. The high precision obtained by our method when the OR prior is used indicates the high quality of the generated skeletons. 

To optimize part positions based on multi-view cues, we generate four label sets $\{(k,s):(3,50),\ (5,10),\ (7,2),\ (7,1)\}$, where the step sizes are in centimeter. We  solve the optimization in four iterations by going from a large and coarse search space towards a small and fine search space, which allows us to more efficiently explore the 3D space. The parameters used in all experiments are $\lambda_1 = 2$ and $ \lambda_2 = 0.5$, that are selected using grid search over a set of 50 frames from the multi-view dataset. The mean and standard deviation (STD) of the 3D Euclidean distances between the predicted body part positions and the ground-truth positions are used to evaluate 3D body part localizations. 

In Table~\ref{tab:localization}, we present the evaluation results for multi-view body part localization as a function of the number of supporting views. Please note that since the head and neck localization errors are close to the  expected error in low-cost RGB-D cameras, we do not update these two joints during our optimization. This table presents localization errors for the initial 3D skeletons obtained by the fusion algorithm and the error after performing the multi-view optimization. One can notice that the proposed multi-view fusion method correctly associates skeletons across views by consistently reducing the localization errors as the number of supporting views increases. However, we observe that if we ignore the left and right labels of the detections and assign the label based on shoulder distances with ground truth, the localization errors decrease by ${\sim}10$ cm for skeletons with one or two supporting views and ${\sim}3$ cm for skeletons with three supporting views. These results indicate that the side detection in individual views is not very reliable. But, if a person is detected in all views, the proposed voting algorithm can make a more reliable prediction. The multi-view optimization significantly reduces the localization error for skeletons with any number of supporting views. Interestingly, the optimization improves the results even for skeletons with one supporting view by properly incorporating the depth-based reprojection costs and detection confidences. The deep RGB-D part detector is the main driver of the optimization. To evaluate the effect of the depth-based reprojection cost, we also report the results without this term in column \lq opt.-$\Phi^{depth}$\rq. The drop in performance highlights its importance. The 2D projections of 3D poses obtained using the proposed multi-view optimization are shown for a few frames in Figure \ref{fig:qualitativeRes}. 

\section{Conclusions}

In this paper, we propose a multi-view RGB-D approach for detecting and estimating the body part positions of medical staff in 3D. A ConvNet-based body part detector  combined with a 3D pairwise deformation model is used to recover body poses in each view. A method based on multi-layer random forests is then proposed to automatically learn {\it a priori}  information about the OR and remove spurious detections per view, which allows us to reliably detect the body poses of persons in the scene. Then, these detections are back-projected to 3D and merged across views. Finally, a novel optimization function is introduced to update the part positions by relying jointly on the body part confidence maps, depth data and multi-view cues. The method has been quantitatively evaluated on a new multi-view dataset acquired during live surgeries. Experimental results show significant improvements over state-of-the-art methods for the task of single-view pose estimation in multi-person scenarios, indicating the benefit of combining deep part detectors and 3D pairwise constraints in building robust models. The multi-view formulation also achieves very promising results showing the benefits of the deep ConvNet detector and of depth data for correctly driving parts towards their optimal locations.  
To the best of our knowledge, this is the first multi-view approach  that performs both human detection and pose estimation in a real scenario without any prior knowledge on the number of persons present, as well as the first multi-view RGB-D approach presented for pose estimation. 

\section*{Acknowledgements}

This work was supported by French state funds managed by the ANR within the Investissements d'Avenir program under references ANR-11-LABX-0004 (Labex CAMI), ANR-10-IDEX-0002-02 (IdEx Unistra), ANR-10-IAHU-02 (IHU Strasbourg) and ANR-11-INBS-0006 (FLI).



{\small
\bibliographystyle{ieee}
\bibliography{refs}
}

\end{document}